\documentclass[10pt,twoside,twocolumn]{IEEEtran}
\pdfoutput=1
\usepackage{cite}
\usepackage{graphicx}
\usepackage{amsmath, amssymb}
\usepackage{array}
\usepackage{siunitx}
\usepackage{float}
\usepackage{algorithm}
\usepackage{algorithmic}
\usepackage{bbold}
\usepackage{ifpdf}
\usepackage{color,soul}
\usepackage{dblfloatfix}

\newcommand{\D}{\mathrm{d}}
\newtheorem{theorem}{Theorem}[section]

\ifCLASSOPTIONcompsoc
  \usepackage[caption=false,font=normalsize,labelfont=sf,textfont=sf]{subfig}
\else
  \usepackage[caption=false,font=footnotesize]{subfig}
\fi
\usepackage{dblfloatfix}
\usepackage{url}
\begin{document}
\title{Designing Color Filters that Make Cameras More Colorimetric}

\author{Graham~D.~Finlayson (g.finlayson@uea.ac.uk)
and~Yuteng~Zhu* (yuteng.zhu@uea.ac.uk) \\
(co-first authors) \\
\thanks{Graham D. Finlayson and Yuteng Zhu are with the School of Computing Sciences, University of East Anglia, NR4 7TJ, Norwich, UK (g.finlayson@uea.ac.uk, corresponding author: yuteng.zhu@uea.ac.uk).}}


\maketitle

\begin{abstract}
When we place a colored filter in front of a camera the effective camera response functions are equal to the given camera spectral sensitivities multiplied by the filter spectral transmittance. In this paper, we solve for the filter which returns the modified sensitivities as close to being a linear transformation from the color matching functions of human visual system as possible.  When this linearity condition - sometimes called the {\it Luther condition} - is approximately met,  the  `camera+filter' system can be used for accurate color measurement. Then, we reformulate our filter design optimisation for making the sensor responses as close to  the CIEXYZ tristimulus values as possible given the knowledge of real measured surfaces and illuminants spectra data. 
This data-driven method in turn is extended to incorporate constraints on the filter (smoothness and bounded transmission). Also, because how  the optimisation is initialised is shown to impact on the performance of the solved-for filters, a multi-initialisation optimisation is developed. 

Experiments demonstrate that, by taking pictures through our optimised color filters we can make cameras significantly more colorimetric.

\end{abstract}

\ifCLASSOPTIONpeerreview
 \begin{center} \bfseries EDICS Category: ELI-COL, ELI-SDP \end{center}
 \fi
\IEEEpeerreviewmaketitle

\section{Introduction}

\IEEEPARstart{D}{igital} cameras are designed in analogy to the trichromatic human visual system which has three cone sensors. If a camera is to capture colors like a human observer, arguably the camera sensors should equal the  cone fundamentals~\cite{hunt2011measuring}. Practically, however, engineering cameras having spectral sensitivities similar to the cone fundamentals is only required  if we wished to construct a biologically plausible model of how we see\cite{wandell1995}.  For most practical applications - e.g.\ photography and video - it is more important that we can transform the recorded device RGBs to drive a display so that the image captured by a camera either looks the same to a human observer or records triplets of numbers - e.g.\ CIE XYZ coordinates - that are referenced to the human visual system\cite{ohta2006colorimetry}. We say that a digital camera is {\it colorimetric} if it meets the so-called Luther condition~\cite{ives1915transformation,luther1927gebiet}, i.e.\ its spectral sensitivity functions are linearly related to the CIE XYZ color matching functions (CMFs).

The Luther condition places a very strong constraint on the shape of camera spectral sensitivities. A strong constraint is required  because the Luther condition effectively assumes that any and all spectral stimuli are possible. However, many studies have shown that the actual spectra (measured in the real world) are far from being arbitrary. Indeed, reflectance spectra tend to be quite smooth\cite{ parkkinen1989,vrhel1994measurement,chiao2000color} and as a consequence can be fit with low dimensional linear basis\cite{maloney1986evaluation,marimont1992linear}. Indeed, spectral basis with dimensions from six to eight, for different applications, are often proposed as adequate models of spectral reflectance. Illuminants by contrast are much less describable by small parameter models. Indeed, for artificial lights such as fluorescent and LED lights, the light spectrum can be very spiky and the number and position of the spikes can vary considerably. And yet, illumninants are also far from being arbitrary. They are designed to have colors near the Planckian locus\cite{ohta2006colorimetry}, a requirement to score highly on color rendering indices \cite{boyce2014lighting}.

Possibly, a more practically useful variant of the Luther condition would be one that is data-driven. That is, where camera RGBs can be mapped to XYZs for the spectral data that are {\it likely to be encountered} in the real world. Equally, in principle, we might consider whether a non-linear mapping could or indeed, should be used.

\begin{figure*}
   \centering
    \includegraphics[width=0.65\textwidth]{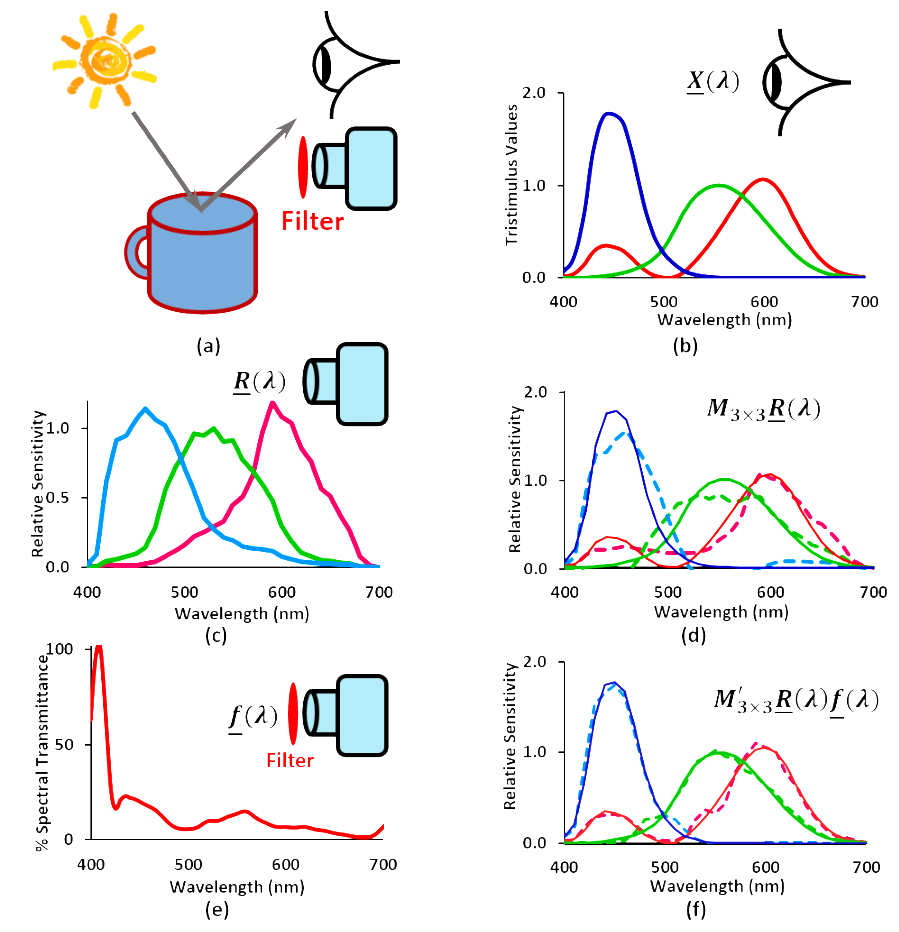}
   \caption{(a) Taking an image of a scene (camera and human eye), (b) the CIE1931 XYZ color matching functions (CMFs) and (c) the spectral sensitivities of a Canon 5D Mark II camera. In (d) we show the least-squares fit of the camera curves to XYZ CMFs. With a specially designed color filter shown in (e), the least-squares fit - of the camera spectral sensitivity curves multiplied by the filter - in (f) is much closer to the XYZ CMFs.}
  \label{fig:pb_statement}
\end{figure*}

It is a classical result \cite{drew1992natural} that if reflectances were {\it exactly} modelled by a 3-dimensional linear model then for a given light spectrum there would be a specific $3 \times 3$ transform matrix  taking camera RGBs to XYZs. While reflectance spectra are not adequately described by a 3-dimensional model, RGBs can be approximately mapped to corresponding XYZs using a $3 \times 3$ matrix.  Indeed, this regression approach is adopted in almost all cameras with good results (we are mostly happy with the colors a camera records). But, as we shall see later the `fit' is not sufficient from a color measurement point of view. 

Of course, rather than using a linear matrix to  map  RGBs to XYZs, we could  use a non-linear transform instead. Possible non-linear methods include Polynomial and Root-polynomial  regressions  and Look-up-tables~\cite{hong2001study,finlayson2015color,hung1993colorimetric}.
However, the linear transform method of using a $3 \times 3$ matrix - even though it is not optimal in terms of fitting error -  has two  advantages compared to most non-linear methods.  First,  the transform scales linearly with exposure. If the scene is made twice as bright (e.g.\ by doubling the quantity of incoming light), the same matrix correctly maps the camera measurements to XYZs (because the magnitude of camera RGBs and  XYZs both double). Typically, non-linear methods do not have this exposure-invariant property (one exception is \cite{andersen2016weighted}).

The second advantage is that a linear transform is, well, linear. The human eye measures color stimuli linearly: at the cone quantal catch level, the response to the sum of two spectral stimuli is equal to the sum of the responses to the two stimuli viewed individually~\cite{wandell1995}. This can be an important {\it physical} consideration. As an example, when we view a surface that has highlights, the recorded color is a convex sum of the so-called {\it body} color (the color name we would assign to the object) and the color of the highlight\cite{klinker1990shafer}, sometimes called the {\it interface} color. If, we are viewing a red shiny plastic surface the body color is red and if the viewing light is white then the interface color is also white (i.e.\ the same as the color of the light). As we move from pure body to pure highlight color, the measured XYZs lie on a 2D plane in the color space. Equally the camera, which at the sensor level has a linear response, will also make measurements that lie on a 2D plane. But, a non-linear correction will distort the plane and the result will be an image that is not physically accurate or even physically plausible. This problem is discussed in detail in~\cite{mackiewicz2016method}.

Practically, the closer the spectral sensitivities of a camera are to being linearly related to the XYZ color matching functions the better it will perform as a tool for color measurement, i.e.\ the more {\it colorimetric} it will become. Interestingly, when we linearly regress RGBs to XYZ tristimuli, we can interpret this as linearly transforming the sensors themselves. That is, a new camera whose sensitivities are modified by the linear regression transform approximately measures the desired XYZs. It follows that one strategy to improving the color measurement capability of a camera would be to change the camera sensitivities (to ones that are more linearly related to the XYZ color matching functions). However,  there are constraints on sensitivities achievable in physically realisable cameras which means we can never get 100\% colorimetric accuracy. 

In this paper, we make an easy modification to the camera spectral sensitivities. We propose simply to place a specially designed filter in front of a camera with the goal of making the filtered RGB measurements it records are as linearly related to the target XYZs as possible.

How we design the {\it best} filter to make a camera colorimetric is the central concern of this paper. 
We develop optimisation methods to solve for the optimal filter that either makes the camera best satisfy the Luther condition or - in a data-driven approach - can best predict measured XYZs for a range of real reflectances and lights. In the rest of this paper we respectively discuss {\bf Luther-condition} and {\bf Data-driven} filters.

The methods we develop are not {\it closed-form} but adopt the alternating least-squares paradigm\cite{zhang2001rank,finlayson2015alternating,Finlayson2019ColorHT}. For the Luther-condition filter optimisation problem we aim to find a filter so that  the filtered camera response functions are a linear transform from the XYZ CMFs~\cite{zhuCIC2018}. In the first step, we find the filter that best maps the spectral sensitivities of the camera to the XYZ CMFs {\it directly}. Then we find the best linear combination of the filtered camera sensitivities that approximate the XYZ sensitivities. Holding this mapping fixed we can solve for a new  best filter. Then we hold the new filter fixed to solve for the best linear transform. We iterate in this way until the procedure converges \cite{golub2012matrix}. Each individual step in the optimisation can be solved, in closed-form, using simple linear least-squares.
In the Data-driven approach we analogously find the filter based on actual measured RGBs and XYZs following the alternating least-squares technique~\cite{zhu2019CCIW}. For the Luther- and Data-driven techniques, the constraint that the recovered filters must be positive is considered.

\begin{figure*}[!ht]
    \centering
        \includegraphics[width=0.75\textwidth]{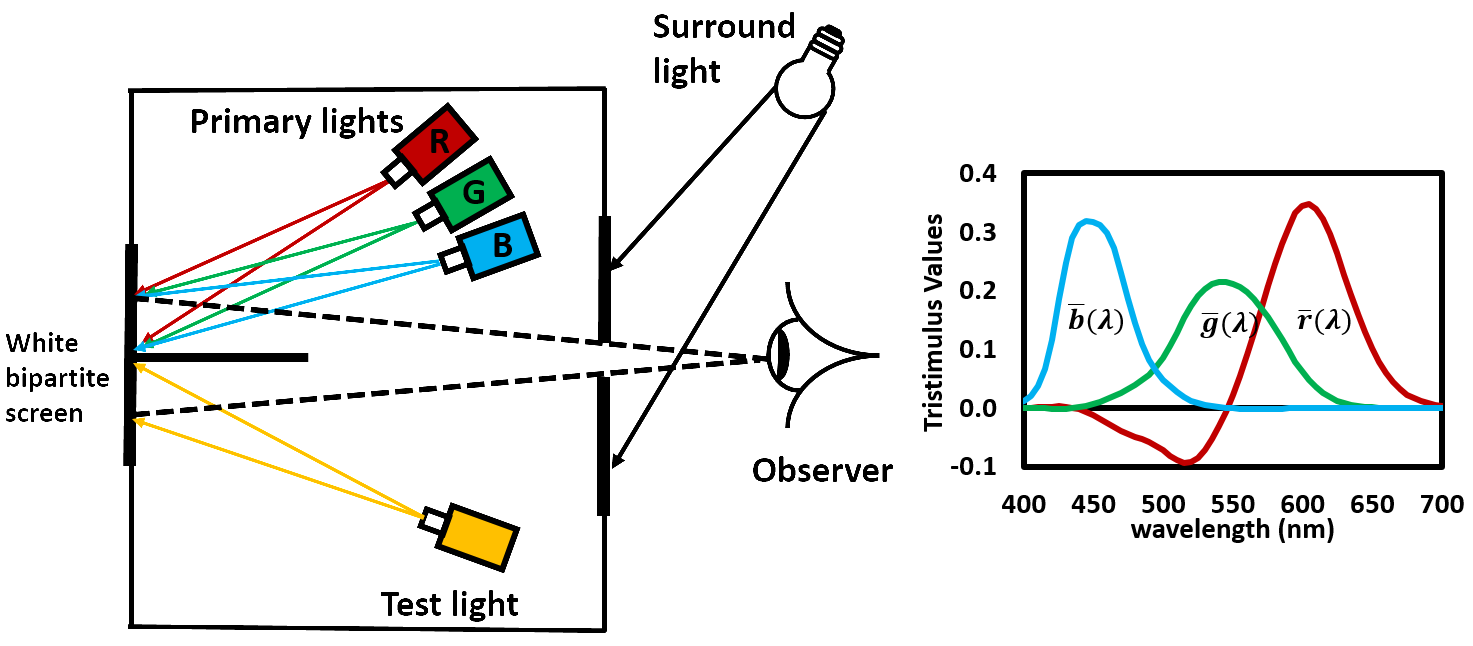}
    \caption{A typical setup that can be used for the color matching experiment (left). 
    Right shows the CIE 1931 RGB color matching functions obtained from the color matching experiments using real R, G, and B primary stimuli.}
    \label{fig:colormatchingexp}
\end{figure*}

Clearly, the filter shown in Fig. 1e is not desireable. In the short wavelengths there is a sharp change in transmittance and as a whole the filter is not smooth. For most of the wavelength range the filter transmits little light ($<$20\%).
Thus, we extend our optimisation framework to incorporate minimum and maximum bounds on transmittance and also that the filters are smooth~\cite{zhu2020EI}.

Experiments demonstrate that  we can find Luther-condition and Data-driven  filters that can dramatically increase the colorimetric accuracy across a large set of commercial cameras.

The rest of the paper is structured as follows. In Section II we review the color matching and color image formation, both ideas underpin our filter design method. The mathematical optimisations for the optimal color filter for a given camera are presented in Section III. The experimental results are reported in Section IV. The paper concludes in Section V.

\section{Background}

\subsection{Color Matching Functions}\label{section:cme}
Color matching functions  provide a quantitative link between the physical light stimuli and the colors perceived by the human vision system. 
Figure~\ref{fig:colormatchingexp} shows a typical  setup for the color matching experiment. The observer views a bipartite field where one side is lit by a test light while the other side is lit by the light mixtures of three primaries (i.e.\ monochromatic red, green, blue lights). The intensities of three primary lights are adjusted  by the observer  to make a visual match, i.e.\ the two stimuli on each side  the bipartite field {\it match} if they look visually indistinguishable  to the observer. Sometimes, no match is possible. In this case one of the primary lights should be added to the test light field. Mathematically, we can model this as if we were subtracting some of the primary light. See \cite{guild1931colorimetric} for more discussion.

By successively setting the test light to be a unit power monochromatic light at sample points across the visible spectrum, we can measure the Color Matching Functions\cite{wyszecki1982color}. That is we find the  RGB mixture that affords a match on a per wavelength basis. Because color matching is linear (the sum of two test spectral lights is matched by the sum of their individual matches) then given the color matching functions we can {\it compute} the match for any arbitrary test light~\cite{krantz1975color}. We simply integrate any test light spectrum with the Color Matching Functions. It can also be shown that the color matching functions are necessarily linearly related to the cone sensitivities~\cite{DeMarco92,stockman2000spectral}. 

Assuming monochromatic primary lights at 650nm, 530nm and 460nm, the resulting CIE RGB CMFs are shown to the right of Fig.~\ref{fig:colormatchingexp}.
The XYZ CMFs are a linear combination of the RGB CMFs, see in Fig. 1b. That XYZ CMFs are used (as opposed to RGB CMFs), in part, because they have, by design, no negative sensitivities. Standardised in 1931, the lack of negatives made pencil and paper calculations with matching curves easy. 

The X, Y and Z scalar values we compute when we integrate a test light spectrum with the XYZ CMFs are called XYZ tristimuli.
In this paper we are interested in using a camera to measure XYZ tristimuli.

The color matching experiment is of direct practical importance. Indeed, suppose a display device can output three colors equal to the R, G and B stimuli used in the color matching experiment. It follows that a camera that had sensitivities equal to RGB Color Matching Functions would {\it see} the correct RGBs to drive the display that would result in a perceptual match to the test light.  Equally, it would suffice that camera's sensitivities were linearly related to the CMFs since we could linearly transform the camera measurements to the correct RGB to drive the display.  Of course, we note that - unlike the RGB color matching functions - we can only {\it drive} the putative display with positive numbers. Consequently, there are  real world colors that cannot be reproduced on displays. 

\subsection{Continuous Color Image Formation}\label{sec:imageformation}

The color of a pixel recorded by a digital camera  depends on three physical factors: the spectral power distribution of the illuminant, the spectral reflectance of the object, and the sensor's spectral sensitivities. The color formation at a pixel, under the Lambertian surface model, can be modeled as
    \begin{equation}
        \label{eq:pixelformation}
    	\rho_{k} = \int_{\omega} \!E(\lambda) S(\lambda)Q_{k}(\lambda)\,\D\lambda \,, \quad k \in \{R, G, B\}
    \end{equation}
where the subscript $k$ indicates the color channel, $\lambda$ denotes wavelength variable defined on the visible spectrum  $\omega$ (approximately, 400nm to 700nm). The functions  $E(\lambda)$ and $S(\lambda)$ respectively denote the spectral power distribution of the illuminant and the spectral reflectance. The function  $Q_{k}(\lambda)$ is the spectral sensitivity of the $k$th camera color channel.

\noindent
Let us define the color signal $C(\lambda)$:

\begin{equation}
    C(\lambda) = E(\lambda)S(\lambda).
    \label{eq:colorsignal}
\end{equation}
Substituting Equation~(\ref{eq:colorsignal}) into Equation~(\ref{eq:pixelformation}):
    \begin{equation}
        \label{eq:pixelreformation}
    	\rho_{k} = \int_{\omega}\! C(\lambda)Q_{k}(\lambda)\,\D\lambda \,, \quad k 		\in \{R, G, B\}.
    \end{equation}
    
\subsection{Discrete Color Image Formation}

In the optimisations that will be presented in Section III, it is useful to recast the continuous integrated responses using discrete representations where spectra are represented as sampled vectors of measurements. Typically and justifiably from a color accuracy point of view\cite{smith1992numerical}, a spectrum can be represented by 31  measurements made between 400nm and 700nm by 10nm sampling interval. Note, the methods we set forth - since they are designed for the discrete domain - are agnostic about the sampling interval. If the data is given at a 5nm sampling distance then each spectrum would be represented as a 61-component vector. Henceforth,  we will talk about spectra being 31-vectors (and this corresponds to the format of most available measured spectral data).

Equation~(\ref{eq:pixelreformation}) is now, equivalently, written as:
\begin{equation}
  \rho_k= \underline{C} \cdot\underline{Q}_k,\quad k \in \{R, G, B\}.
  \label{eq:Graham1}
\end{equation}
Respectively $\underline{C}$ and $\underline{Q}_k$ denote the sampled version of the color signal spectrum and the $k$th spectral sensitivity function. We assume the sampling distance is incorporated in the spectral sensitivity vectors. Here `$\cdot$' denotes the dot product of two vectors.

One advantage of transcribing image formation into vector notation is that we can usefully deploy the tools of linear algebra. Let the  matrices $C$ and $Q$ denote $31\times n$ and $31\times 3$ matrices whose columns respectively contain $n$ color signal spectra and the 3 device spectral sensitivities. The $n\times 3$ set of RGB responses can be written as a single concise expression:

\begin{equation}
  P = C^{T}Q,\quad  	
  \label{eq:P}
\end{equation}
where $^T$ denotes the matrix transpose.

Denoting, {\it X} as a $31\times 3$ matrix whose colmuns contain the discrete XYZ color matiching functions, the  XYZ tristimulus responses can be written as: 
\begin{equation}
  X = C^{T}\text{\it X}.	
  \label{eq:X}
\end{equation}

\subsection{The Luther Condition for Camera Spectral Sensitivities}

Let us denote individual camera and XYZ responses as $\underline{\rho}$ and $\underline{x}$. We can use a camera to {\it exactly} measure colors if and only if there exists a function such that $g(\underline{\rho})=\underline{x}$ for all spectra. It follows that if a pair of spectra  integrates to  the same RGB then this pair must also integrate to the same XYZ:

\begin{equation}
 \underline{C}^T_1Q= \underline{C}^T_2Q \;\Rightarrow \; \underline{C}^T_1\text{\it X}= \underline{C}^T_2\text{\it X}
  \label{eq:Y}
\end{equation}
 
This implies $\underline{C}^T_1-\underline{C}^T_2$ is simultaneously in the null space of $Q$ and {\it X}. Since any spectrum in the null-space of $Q$ is a physically plausible spectral difference this implies that the null-spaces of $Q$ and {\it X} are the same and this in turn implies the Luther condition 

\begin{equation}
 {\it X}=QM 
 \label{eq:Luter1}
 \end{equation}

 \noindent
 where $M$ is a $3\times 3$ matrix. See \cite{HORN1984} for the original proof of the Luther condition (which we precis above). 
 
 \subsection{The Data-driven Luther Condition }
 
The Luther condition for Spectral Sensitivities  presupposes that any and all physical spectra are plausible. However, we know that reflectance spectra are smooth\cite{chiao2000color}. Lights while more arbitrary are designed to integrate to fall near or close to the Planckian locus\cite{ohta2006colorimetry} and to score well on measures such as the Color Rendering Index\cite{boyce2014lighting}. Pragmatically, a camera is colorimetric if its responses are a linear transformation from the XYZ tristimulus values
\begin{equation}
C^T\text{\it X} = C^TQM
  \label{eq:YY}
\end{equation}
where $M$ is a $3\times 3$ matrix.

\subsection{Filter Design}

There are many papers in the literature where the spectral sensitivities that a camera {\it should have} had are designed~\cite{vora1997design,vrhel1994filter,vrhel1995optimal,wolski1996opt,vora1997analysis}. For example, we can solve for the camera spectral sensitivities with respect to which the RGBs mapped with respect to many illuminant spectra can be mapped to corresponding XYZs for a single target illuminant. The procedure where we  measure colors under a changing light - e.g.\ in the real world -  but then reference (that is, map) these colors back to a fixed illuminant viewing condition is a standard methodology in color science. Curiously, the best sensors that solve this problem are not linearly related to the XYZ CMFs \cite{sharma1997digital}.

Perhaps, the closest work to our study is~\cite{farrell1995method}. Here two images are captured. The first with the native spectral sensitivities and the second through a colored filter. The emphasis of that work was to increase the spectral dimensionality of the capture device.  Since we make two captures we have 6 measurements per pixel. Effectively we have a 6-dimensional sensor set. We match target XYZs by applying a $6\times 3$ correction matrix. In~\cite{farrell1995method}, the best filter was  chosen from a  set of commercially available filters. 

There are many other works which propose recovering spectral information by capturing multiple exposures of the same scene through multiple filters, e.g.\ \cite{hardeberg2001acquisition,connah2001recovering,nieves2005multispectral,ng2006subspace,valero2007recovering}. A disadvantage of all these methods is that the capture process is longer and more laborious. Filters need to be changed between exposures. The multi-exposure process then need to be registered. Image registration remains a far from solved problem.  Moreover, scene elements between exposures may move (making registration impossible). 

The method we propose here is much simpler. We simply place a specially designed filter in front of a camera and then take conventional single exposure images.

\section{Designing a Filter to Make a Camera more Colorimetric} \label{sec:method}

\subsection{Luther-condition Filters}

The Luther condition states that a camera system is colorimetric if its sensitivities are a linear transform from the XYZ color matching functions. We propose a modified  Luther condition where a camera is said to be colorimetric if there exists a physically realisable filter which, when placed in front of the camera, generates {\it effective} sensitivities which are a linear transform from the XYZ CMFs. 

Physically, the role of a filter, which absorbs light on a per wavelength basis, is multiplicative. If $f(\lambda)$ is a transmittance filter and $\underline{Q}(\lambda)$ denotes the camera sensitivities then $f(\lambda)\underline{Q}(\lambda)$ is a physically accurate model of the effect of placing a filter in front of the camera sensor.

In Equation (\ref{eq:Luther}) we write an optimisation statement for the Filter-based Luther condition:

\begin{equation}
	\min\limits_{\underline{f},M}\parallel{diag(\underline{f})QM - \text{\it X}}\parallel^2_F\;\text{s.t.}\;\underline{f}>0
\label{eq:Luther}
\end{equation}
Here $Q$ and {\it X} are $31 \times 3$ matrices encoding respectively the spectral sensitivities of a digital camera and the CIE standard XYZ color matching curves. The 31-vector $\underline{f}$ is the sampled equivalent of the continuous filter function  $f(\lambda)$. The $diag()$ operation converts a vector into a diagonal matrix (the vector components appear along the diagonal). The meaning of $diag(\underline{f})Q$  is the same as $f(\lambda)\underline{Q}(\lambda)$, i.e. the diagonal matrix allows us to express component-wise multiplication. $M$ denotes a $3\times 3$ matrix. We minimise the square of Frobenius norm $\parallel \, \parallel^2_F$ (we minimise the sum of squares error). Notice the constraint that the filter value is larger than 0  (physically, we cannot have a filter that has negative transmittance).
 
We do not have to constrain the maximum transmittance because
we can only solve for $\underline{f}$ and $M$ up to an unknown scaling factor. Indeed, suppose the filter $\underline{f}$ is returned where the max transmittance is larger than 1. The fitting error in Equation
(\ref{eq:Luther}) is  unchanged if we divide $\underline{f}$ by its maximum value (resulting in a max transmittance of 100\%) so long as we multiply the corresponding correction matrix $M$ by the same value.

We minimise Equation (\ref{eq:Luther})  using an  alternating least-squares (ALS) procedure given in the following algorithm.

\begin{algorithm}[h!]
	\caption{ALS algorithm for Luther-condition optimisation}
	\label{algo1}
	\begin{algorithmic}[1]
		\STATE{$i=0, Q^{0}=Q $}
		\REPEAT
		\STATE{$i = i+1$}
		\STATE{$\min\limits_{\underline{f}^{i}} \parallel diag(\underline{f}^{i})Q^{i-1} - \text{\it X}\parallel_{F}^2$}
		\label{codeD}
		\STATE{$\min\limits_{M^{i}} \parallel diag(\underline{f}^{i})Q^{i-1}M^{i} - \text{\it X}\parallel_{F}^2$}
		\label{codeM}
		\STATE{$Q^{i} = diag(\underline{f}^{i})Q^{i-1}M^{i}$}
		\UNTIL{$\parallel Q^{i} - Q^{i-1} \parallel_{F}^2 \, < \, \epsilon $} \\
		\STATE{$ \underline{f}^{Luther} = \prod_{s=1}^i {\underline{f}^{s}} $\quad and \quad $M = \prod_{s=1}^i {M^{s}} $}	\label{algo1_finalstep}
	\end{algorithmic}
\end{algorithm}

\noindent
where $\prod$ denotes component-wise matrix multiplication. 
Steps 4 and 5 - where we find the filter and then the linear transform - are solved using simple, closed-form least-squares estimation. For completeness we provide details of how these calculations are made in the Appendix.

At each iteration, the filter and linear transform -  $\underline{f^i}$ and $M^i$ - are calculated relative to the previous $i-1$ filters and matrices. It follows in step~\ref{algo1_finalstep} that the final solution is the multiplication of all the per-iteration solutions: $ \underline{f}^{Luther} = \prod_{s=1}^i {\underline{f}^{s}} $ (component-wise vector multiplication) and $M_j = \prod_{s=1}^i {M^{s}}$ (component-wise matrix multiplication).

Notice nowhere in the above procedure do we constrain the filter transmittance to be larger than 0 (even although this constraint is in the optimisation statement). Empirically, we found that the optimised filter is always positive for all the cameras we tested (see experimental section). Moreover, Theorem $3.1$, presented below, proves that when there exists a filter which makes the camera sensors perfectly colorimetric that the filter has to be everywhere positive.

The theorem is presented for continuous spectral sensitivity functions. As such, we write the XYZ CMFs and camera sensitivities as vector functions: $\underline{\text{\it X}}(\lambda)$ and $\underline{Q}(\lambda)$. In this representation, we, effectively, have taken transposes of the matrices $Q$ and $\text{\it X}$. So, here, we write $M^T$ for the $3 \times 3$ matrix. Of course, matrix $M$ in the proof and in the algorithm presented above signifies the same linear transform.

\vspace{0.1in}

\begin{theorem}
Assuming there exists an exact solution that $f(\lambda)>0$ for $M^T\underline{Q}(\lambda)f(\lambda)=\underline{\text{\it X}}(\lambda)$, the variable $\lambda$ is defined over the domain where  $\underline{\text{\it X}}(\lambda)>0$ are continuous  and full rank (no one spectral  sensitivity  can be written as a linear combination of the other two) and $\underline{Q}(\lambda)$ are also continuous then  $f(\lambda)>0$.
\end{theorem}

\vspace{0.1in}
\noindent

 {\it Proof}: First we remark on the continuity of the camera and XYZ functions. Both are the result of physical processes which are continuous in nature. To our knowledge it is not possible to make a physical sensor system that captures light which has discontinuous sensitivities. And, in terms of physiological systems, biological sensor response functions are always continuous. 
 
 Next, if {$f(\lambda)<0$} across all wavelengths and \hspace{.5in} $\;$ $M^T\underline{Q}(\lambda)f(\lambda)=\underline{\text{\it X}}(\lambda)$, then $-M^T\underline{Q}(\lambda)*(-f(\lambda))=\underline{\text{\it X}}(\lambda)$. In this case $-f(\lambda)$ must be all positive and so an all-positive filter can be found. The interesting case to consider is when the filter has both negative and positive values.

Clearly the $3\times 3$ matrix $M^T$ must be full rank otherwise the mapped camera sensitivities would be rank deficient and therefore could not model the CMFs. Equally,  multiplying by a filter does not change the rank of the sensor set. Because, by assumption $\underline{Q}(\lambda)$ are continuous it follows that $f(\lambda)$ must also be a continuous function since otherwise $M^T\underline{Q}(\lambda)f(\lambda)$ would be discontinuous (multiplying a continuous and discontinuous functions together, save for the case where one of the functions is everywhere 0, results in a discontinuous function).

As $f(\lambda)$ is continuous if the function has both negative and positive values there must be at least one wavelength  $\lambda_v$ where $f(\lambda_v)=0$ and so $M^T\underline{Q}(\lambda_v)f(\lambda_v)=\underline{0}$. But, this cannot be the case since the XYZ color matching functions are not all zero at any given wavelength within the defined domain.

\hspace{3in} QED

\subsection{Data-driven Filters} \label{sec_datadriven}

\noindent
{\it Simple Case:} in the {\it simple} Data-driven approach, we look for a  color filter and the $3 \times 3$ color correction matrix that, in a least-squares sense, best maps camera measurements for a training color signal data set to the corresponding ground-truth XYZ tristimulus values. Denoting a collection of $n$ color signals in the $n\times 31$ color signal matrix $C$,  the Data-driven optimisation is written as:
\begin{equation}
		\min \limits_{\underline{f}, M} \parallel C^{T}diag(\underline{f})QM - C^{T}\text{\it X} \parallel_F^2 \;\text{s.t.}\;\underline{f}>0
	\label{eq:pb_form}
\end{equation}

Solving Equation~(\ref{eq:pb_form}) depends on the structure of the color signal matrix. If we choose $C={\cal I}_{31}$ (the $31\times 31$ identity matrix) then we can solve this optimisation using Algorithm 1 (in this case, Equations~(\ref{eq:pb_form}) and (\ref{eq:Luther}) are the same). This assumption is related to the Maximum Ignorance assumption~\cite{sharma2002digital} where all possible spectra are considered equally likely. 

\vspace{0.1in}
\noindent
{\it General Case:}
Let us develop a more general optimisation statement. One,  where we have $cnt$  color signal matrices - denoted $C_j$ ($j\in\{1,2,\cdots, cnt\})$ and the corresponding $cnt$ color correction matrices $M_j$. Each color signal matrix typically corresponds to a training set of surface reflectances illuminated by a single spectrum of light $\underline{E}_j$, thus the color signal matrix is $C_j=diag(\underline{E}_j)S$, where $S$ is a $31\times n$ matrix of reflectances, one reflectance spectrum per column. But, the different light assumption is not a necessary assumption. As an example, we might mix color signals for the Maximum Ignorance assumption with measured data (i.e. two color signal matrices) where both measurement sets contain multiple lights. 

The general Data-driven filter optimisation problem is written as:

\begin{equation}
\min \limits_{\underline{f}, M_j}	\sum_{j=1}^{cnt}	 \parallel C^{T}_jdiag(\underline{f})QM_j - C^{T}_k\text{\it X} \parallel_F^2  \; \text{s.t.}\;\underline{f}>0
	\label{eq:pb_form2}
\end{equation}

\noindent
and is solved using Algorithm 2.

Finally,  $k$ could denote some other privileged standard reference viewing condition (where the reference viewing illuminant is not in the set of training lights). For example, in color measurement we are often interested in the XYZ tristimuli for a daylight illuminant D65 which has a prescribed but not easily physically replicable spectrum.

\begin{algorithm}[t]
	\caption{ALS algorithm for the Data-driven optimisation}
	\begin{algorithmic}[1]
		\STATE{$i=0, \underline{f}^0=\underline{f}^{seed}, Q^{0}=diag(\underline{f}^0)Q$} \label{algo2_init}
		\REPEAT
		\STATE{$i = i+1$}
      \STATE{$\min\limits_{M^{i}_j} \parallel{C^T_j Q^{i-1}_j M^{i}_j - C^T_k \text{\it X}}\parallel_{F}^2, \; j = 1,2,...,cnt$} \label{codeMij}
  		\label{codeM2}
		\STATE{$\min\limits_{\underline{f}^i} \sum_{j=1}^{cnt} \parallel{C^T_j diag(\underline{f}^i) Q^{i-1}_jM^{i}_j - C^T_k \text{\it X}}\parallel_{F}^2,\;\underline{f}^i>\underline{0}$}
		\label{codeD2}		
		\STATE{$Q^{i}_j=  diag(\underline{f}^{i})Q^{i-1}_j M_{j}^{i}$}
		\UNTIL{$\forall(j) \parallel{Q^{i}_j - Q^{i-1}_j}\parallel_{F}^2 \, < \, \epsilon $} \\
		\STATE{$ \underline{f}^{Data} = \prod_{s=0}^i {\underline{f}^{s}} $\quad and \quad $M_j = \prod_{s=1}^i {M^{s}_j } $	}
	\end{algorithmic}
	\label{algo2}
\end{algorithm}

We are going to solve Equation (\ref{eq:pb_form2}) for the filter $\underline{f}$ using an  alternating least-squares procedure. Notice that the input to the optimisation is an initial filter guess denoted by $\underline{f}^{seed}$. 
Let us consider 3 candidate minimisations corresponding to 3 common scenarios for color measurement each of which can be solved using Algorithm~\ref{algo2}.

\noindent
{\it 1) Multiple Lights:} Here we assume that $j$ indexes over $cnt$ illuminants and $k=j$ (per illuminant we make the target XYZs using the same color signals).  We find a single filter which given per-illuminant  optimal least-square $3\times 3$ correction  matrices will best fit camera data to the corresponding multi-illuminant XYZs.

\noindent
{\it 2) Multiple measurement lights, single target light:} Again $j$ indexes over $cnt$ illuminants. But, the target is a single illuminant, for example CIE D65\cite{ohta2006colorimetry}.

\noindent
{\it 3) Single Light}. This case is, in effect, the simple restriction of the general case, $cnt=1$. We have one measurement light and one target light. Like the Luther-condition optimisation, we solve for a single correction matrix.
 
In Algorithm 2,  it is straightforward to solve for the $j$th color correction matrix at iteration step $i$, $M^i_j$, in step~\ref{codeMij} using the Moore-Penrose inverse. Step~\ref{codeD2}, where we find the filter, can also be solved directly using simple least-squares, although the basic equations need to be rearranged. Details of the least-squares computation are given in the Appendix. Here, to ensure that the filter is all positive we can also solve for the filter by solving the optimisation subject to the positivity constraints, we solve a quadratic programming problem~\cite{luenberger1984linear} (unlike the Luther-condition case there is no a prior physical reason why the best filter should be all positive). 

Quadratic programming allows linear least-squares problem subject to linear constraints to be solved rapidly and, crucially, a global optimum is found.

\subsection{Adding Filter Constraints} \label{sec:method_filterconstrains}

By default the filter found using Algorithm 2 can be arbitrarily non-smooth and might also be very non-transmissive. Non-smoothness limits manufacturability (at least with dye based filters) and a filter that absorbs most of the incoming light would, perforce, have limited practical utility. Both these problems can be solved by placing constraints on the filter optimisation.

Let us now constrain the target filter $\underline{f}$ according to:
\begin{equation}
\underline{f}=B\underline{c},\;\;\text{s.t.}\;\;f_{min}\leq\underline{f}\leq f_{max}
	\label{eq:constraints}
\end{equation}

\noindent
here $B$ denotes a $31\times m$ basis matrix with each column representing a basis and the vector $\underline{c}$ denotes an $m$-component coefficient vector. The scalars $f_{min}$ and $f_{max}$ denote lower and upper thresholds on the desired transmittance  of the filter; specifically, $f_{max}$ is set to 1 by default as fully transmissive and $f_{min}$ is a positive value between 0 and 1. Equation~(\ref{eq:constraints}) forces the optimised filter  to be in the span of the column vectors of  $B$ and to meet the min and max transmittance constraints.
By judicious choice of the basis we can effectively bound the smoothness of the filter. For example, we could choose the first $m$ terms of the discrete cosine transform basis expansion\cite{strang1999discrete}.

With respect to the new filter representation, we rewrite the new overall  filter design optimisation in Equation~(\ref{eq:pb_form2}) as
\begin{equation}
    		\min \limits_{\underline{c}, M_j}\sum_{j=1}^{cnt} \parallel C^{T}_j diag(B\underline{c})QM_j - C^{T}_k\text{\it X} \parallel_F^2,   \;f_{min}\leq  B\underline{c} \leq f_{max}
	\label{eq:pb_form_constraints}
\end{equation}

\noindent
The current minimisation can be solved using the same alternating least-squares paradigm of Algorithm 2. But, in step~\ref{codeD2}, we need to substitute  the constraint $\underline{f}^i>\underline{0}$ with
\begin{equation}
f_{min}\leq diag(\prod_{s=0}^{i-1} \underline{f}^s)\underline{f}^i = B\underline{c}^i \leq f_{max}
\label{eq:optc}
\end{equation}
\noindent
That is, the filter we find at the $i$th iteration when multiplied by all the filters from the previous iterations is constrained to be in the basis $B$. 

Looking at Equation~(\ref{eq:optc}) we see that 
\begin{equation}
\underline{f}^i = [diag(\prod_{s=0}^{i-1} \underline{f}^s)]^{-1}B\underline{c}^i
\label{eq:newf}
\end{equation}

\noindent
That is, effectively the basis for the {\it i}th filter changes at each iteration. Again we can solve step~\ref{codeD2} subject to the constraints of Equation~(\ref{eq:newf}) using Quadratic Programming. 

Finally, we note that we could, of course, rewrite Algorithm 2 so that at each iteration we solve for a filter defined by a coefficient weight directly (we could solve for an $m$-term  coefficient vector rather than a 31-component filter). The two formalisms are equivalent. Here, we chose to solve for the per iteration filter for notational convenience: we can use the general Data-driven algorithm and simply change how we calculate the per iteration filter optimisation.

\subsection{Initialising the  Data-driven Optimisation} \label{subsec:initial}

Alternating least-squares is guaranteed to converge but it will not necessarily return the  global optimal result. But, it is deterministic. So, given the state of the correction matrices and filter at the $i$th iteration we will ultimately arrive at the same solution. Equally, if we change the initialisation condition, $\underline{f}^{seed}$, in Algorithm 2, we may end up solving for a different filter.  Empirically, we observed that the filter returned by the algorithm depends strongly on the initial filter that seeds the optimisation.

Let us consider 3 different ways to seed the Data-driven optimisation:

\noindent
{\it 1) Default:}  $\underline{f}^{seed}= \underline{1}$. This uniform unit vector denotes a fully transmissive filter over the spectrum. \\
\noindent
{\it 2) Luther filter:} $\underline{f}^{seed}=\underline{f}^{Luther}$. That is we seed the Data-driven optimisation with the optimal Luther-condition filter found using Algorithm~\ref{algo1}. \\
\noindent
{\it 3) By sampling:} Here we find a set of sample filters ${\cal F}$ (which meet our smoothness and transmittance boundedness constraints) and for each filter sample, $\underline{f}\in {\cal F}^{seed}$, we will run Algorithm 2.

Algorithm 3  generates $\#filters$ (number of initial filters)---subject to bounded smoothness constraints---by uniformly and randomly sampling the filter coefficient space. Before sampling, the algorithm first finds the min and max values of the coefficients (which are calculated in each of the $m$ dimensions individually).  Explicitly, for the $i$th component in vector $\underline{c}$, we denote its minimum and maximum values as $c_i^{min}$ and $c_i^{max}$.

In Algorithm 3, for the minimum value of the $i$th coefficient, we write: 
$c^{min}_i = \min {c_i} ,\,\text{s.t.} \;	f_{min}\leq B\underline{c} \leq f_{max}, i=1,2,...,m$. That is, over all possible vectors $\underline{c}$, which satisfy the transmittance constraints, we take note of the minimum value of the $i$th component. That is we find the minimum value that $c_i$ can be over the set of all possible solutions. The maximum of the $i$th coefficient term is written similarly (see step 3, Algorithm 3).
The minimum and maximum values of $c_i$ can be solved using linear programming\cite{luenberger1984linear}.

All $m$ min and max components, taken together, make  the two  vectors  $\underline{c}^{min}$ and $\underline{c}^{max}$. These vectors together define the extremal values in each dimension of an m-dimensional hypercube. A vector that lies outside the hypercube is guaranteed not to satisfy the boundedness and transmittance constraints we have placed on our filters. This hypercube usefully delimits our search space (of the sample set of solutions).

\begin{algorithm}[!h]
	\caption{Algorithm for generating an initial filter set}
	\begin{algorithmic}[1]
		\STATE{${\cal F}=\{\}$}
		\STATE{$c^{min}_i = \min {c_i} ,\,\text{s.t.} \;	f_{min}\leq B\underline{c} \leq f_{max}, i=1,2,...,m$} \label{stepmin}
		\STATE{$c^{max}_i = \max {c_i}, \,\text{s.t.} \; f_{min}\leq B\underline{c} \leq f_{max},\; i=1,2,...,m$} \label{stepmax}
       	\WHILE{$cardinality({\cal F}) < \# filters$}
    		\STATE{ ${c}_i \sim \it{U} \left({c}_i^{min},{c}_i^{max} \right ), i=1,2,...,m$}
		\STATE{$\underline{f}=B\underline{c}\;,\;(\underline{c}=[c_1\;c_2\;\cdots \; c_m]^T)$}
		    \IF{$f_{min}\leq \underline{f} \leq f_{max} \; \& \;
		    \{\forall{\underline{q}} \in {\cal F}: angle(\underline{f},\underline{q}) > {\theta}$\}} \label{vectorAngle}
		        \STATE{${\cal F} \gets {\cal F} \cup \{\underline{f}\}$}
		    \ENDIF
      \ENDWHILE \\
	\end{algorithmic}
	\label{algo3}
\end{algorithm}

To generate a set of filters for initialising the optimisation (solved in Algorithm 2) we will sample uniformly and randomly this hypercube.  We use the notation  ${c}_i \sim \it{U} \left({c}^{min}_i,{c}^{max}_i \right )$  to denote sampling a number in the interval $[c^{min}_i,c^{max}_i]$ uniformly and randomly.  A filter constructed from the corresponding vector $\underline{c}=[c_1\;c_2\;\cdots\;c_m]^T$ ($\underline{f}=B\underline{c})$, will be added into the initial filter set ${\cal F}$ only if it lies within the transmittance bounds {\bf and} it is sufficiently far from those filters already in the set, see step 7. In algorithm 3 {\it sufficiently far} means at least $\theta$ degrees from the other set members. The function $cardinality()$ returns the number of members in a set.

\section{Results}

\subsection{Experiments for Luther-condition Optimised Filters} \label{sec:LuthResults}

Let us return to the example shown in Fig.~1.  The optimal Luther-condition filter solved using Algorithm~\ref{algo1} is shown in Fig.~1e. We multiply the camera sensors by this filter and then find the linear least-squares transform mapping the new effective sensitivities to the XYZ matching functions. The fitted filtered camera sensitivities (to the XYZ color matching functions) are shown  Fig.~1f. In contrast, Fig. 1d, shows the native camera spectral sensitivities fitted to the XYZs. Visually, the addition of our derived filter makes the camera much more colorimetric. 

The reader will notice that the filter, Fig.~1e, absorbs more than 80\% of the light  except at the shortest wavelengths where it is maximally transmissive. This need not be a problem for color measurement as we can increase exposure time, for example. Though, it does mean that the camera with and without the filter  would, for the same recording conditions, capture significantly less light. And, this could result in an increase in the  noise in the final image output by a camera reproduction pipeline. Indeed, if we deploy this filter - and keep the capture conditions the same - we would need to `scale up' the recorded values and this operation also scales up the noise. Effectively, we capture an image at a higher ISO number.

\subsection{Vora-Values}

To quantitatively measure the {\it spectral} match between the filtered and linear transformed camera sensors and the XYZ color matching functions, we calculate the Vora-Value\cite{vora1993measure}.
The Vora-Value measures the  closeness between the spaces spanned by a set of filter sensitivities set ${Q}$  and that by the color matching functions \text{\it X}. It is defined as 
\begin{equation}
	\nu(\text{\it X},{Q}) =
	\frac{Trace({ QQ}^{+}\text{\it XX}^+)}{3}
\label{eq:VV}
\end{equation}
\noindent
where $Trace()$ returns the sum of the diagonal of a matrix and $^+$ is the Moore-Penrose inverse (see Appendix). The Vora-Value is a number between 0 and 1 where 1 indicates the two sensitivity sets span the same space, i.e.\ the Luther condition is fully satisfied. While there is not a guide on what different Vora-Values {\it mean}, empirically we have found when the Vora-Value is respectively larger than 0.95 and 0.99 then we have acceptable and very good color measurement performance. An explanation of why the Vora-Value is useful for quantifying the color measurement potential of a set of color filters together with its derivation  can be found in~\cite{vora1993measure}.

\begin{figure}[!t]%
    \centering
      \includegraphics[width=0.48\textwidth]{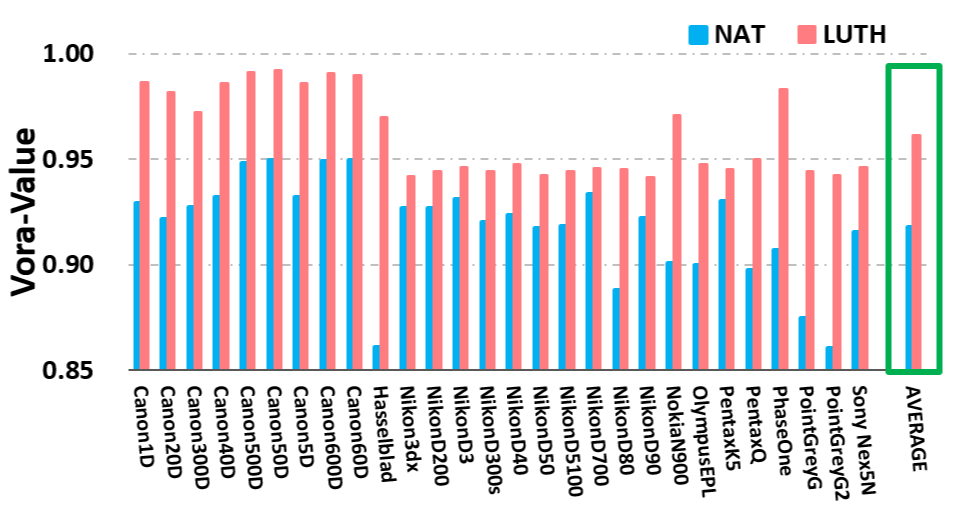}
    \caption{Spectral match to Luther-condition by (a) unfiltered {\bf NAT}ive sensitivities and (b) filtered {\bf LUTH}er-condition optimised sensitivities for a group of 28 cameras in terms of Vora-Values. 
    The overall average Vora-Values of the camera group by both conditions are shown in the green box. }
    \label{fig:vv}
\end{figure}

The  Vora-Value performance for a set of 28  camera spectral sensitivities \cite{jiang2013space}  - with and without  their optimised filters - are shown in Fig.~\ref{fig:vv}.  The Vora-Value for the unfiltered, {\bf NAT}ive sensitivities are shown in blue and for the {\bf LUTH}er-condition optimised sensitivities in red. On average, the native Vora-Value  is 0.918 but when the optimised filter is added it increases to 0.961. This digital cameras data set \cite{jiang2013space} comprises of diverse camera types, including professional DSLRs, point-and-shoot, industrial and mobile cameras.

Note, we make a distinction between using a camera for color measurement and for making attractive looking images. Here, we are interested in using a camera to measure XYZ tristimuli (or measures like CIE Lab values that are derived from tristimuli~\cite{ohta2006colorimetry}). From a measurement perspective, we need higher tolerances and a higher Vora-Value. Clearly, from the point of view of making attractive images cameras that have Vora-Values less than 0.95 can work very well. Indeed, many of the 28 cameras with Vora-Values less than 0.95 can still take images that look appealing from a photographic perspective. But, commercial cameras are not suitable vehicles for accurate color measurement. Quantitative color measurement results are presented in the next section. 

\subsection{Color Measurement Experiments}

Now let us evaluate the derived Luther-condition and Data-driven filters in terms of a perceptually relevant {\it color measurement/image reproduction} metric. The CIELAB $\Delta E_{ab}^*$ \cite{ohta2006colorimetry} - the Euclidean distance calculated between two tristimuli - is  a single number that roughly correlates with  the perceived perceptual distance  between two colors. One  $\Delta E_{ab}^*$ corresponds approximately to a `Just Noticeable Difference' to a human observer~\cite{wyszecki1982color}. When $\Delta E_{ab}^*$ is less than 1 we say that the difference is imperceptible to the visual system. 

\noindent

\subsubsection{Single Light Case}

Let us use the  Canon 5D Mark II camera spectral sensitivities as a putative measurement device and quantify how well it can measure colors - with and without a filter.
In this experiment  the measurement light is either
 a CIE D65 (bluish) or a  CIE A (yellowish) illuminant.   For reflectances we use the SFU set of 1995 spectra \cite{barnard2002data} (itself a composite of many widely used sets).
 The 1995 XYZs for these reflectance and lights are the {\it ground-truth} with respect to which we measure color measurement error. 
 
 Using the Canon camera sensitivities, the reflectance spectra and either the CIE D65 and A lights we numerically calculate two sets of 1995 RGBs. 
 Now, we linearly regress the  RGBs for each light to their corresponding ground-truths (we map the native RGBs for CIE D65 and A to respectively the XYZs under the same lights). We call these color corrected RGBs the {\bf NAT}ive camera predictions (and we adopt this notation in the results shown in Table~\ref{tab:Canon_de_all}). Rows 1 and 4 of Table~\ref{tab:Canon_de_all} report the mean, median, 90, 95 and 99 percentile and the maximum CIELAB $\Delta E$  error for CIE D65 and CIE A lights.
 
 Now let us  place a filter in front of the camera. 
 Again, we calculate two sets of RGBs (one for each light) for the camera spectral sensitivities multiplied  by the filter found using the Luther-condition optimisation (Algorithm 1). The recorded filtered RGBs are mapped  best  to corresponding XYZs using linear regression. The {\bf LUTH}er $\Delta E_{ab}^*$  error statistics are shown in rows 2 and 5. It is clear that placing a Luther-condition Filter substantially increases the ability of the camera to measure colors accurately. Across all metrics the $\Delta E_{ab}^*$  errors reported are about a third of those found when a filter is not used.
 
 We repeat the experiment for a Data-driven color filter (found using Algorithm 2 where the seed filter for the optimisation is the Luther-condition optimised filter).  Again, the two sets of filtered RGBs are linearly mapped to corresponding XYZs to minimise a least-squares error. Results for the corrected {\bf DATA}-driven filtered RGBs for the two lights are reported in rows 3 and 6 of Table I. Clearly, the camera plus filter can measure colors more accurately compared to the case where a filter is not used. Across all metrics the $\Delta E_{ab}^*$ errors reported are about a quarter of those found when a filter is not used.
 
 Significantly,  the best Data-driven filter also delivers improved performance compared with the results reported for the  Luther-condition optimised filter. The errors are further reduced by about a quarter. Incorporating knowledge of typical lights and surfaces into the optimisation leads to improved color measurement performance.

 \subsubsection{Multiple Lights Case}

We now repeat the experiment for a  set of 102 measured lights\cite{barnard2002data}. The results of this second experiment are  shown in the last 3 rows of Table~\ref{tab:Canon_de_all}. Here, the reported error statistics are averages. For each illuminant - as described in the single light case above -  we calculate the mean, median, 90 percentile, 95 percentile, 99 percentile and maximum $\Delta E_{ab}^*$. That is, we calculate 6 error measures for 102 lights. We then take the mean of each error statistic over all the lights. The aggregate illuminant set performance is reported in rows 7, 8 and 9 of Table I for respectively unfiltered RGBs and RGBs measured with respect to Luther-condition and Data-driven optimised filters.

In terms of the reported errors of the raw RGBs compared to the filtered RGBs for the Luther- and Data-driven filters we see the same data trend for the multiple lights case as we saw previously for single lights. A Luther-condition derived filter reduces the measurement error by 2/3 and for the Data-driven filter the measurement error is reduced by 3/4, on average.

\subsubsection{Multiple Cameras}

\begin{figure}[!tb]%
    \centering
      \subfloat[mean $\Delta E_{ab}^*$ color error]{\includegraphics[width=0.48\textwidth]{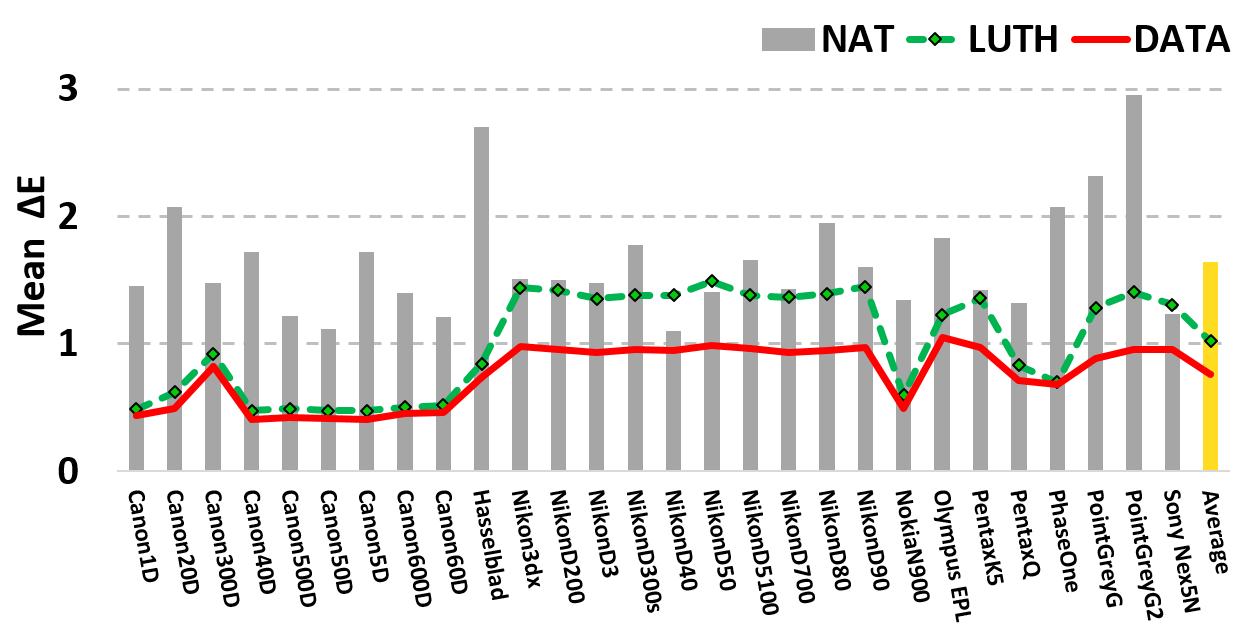} \label{fig:de_mean_28cam} }
	\hfill
	 \subfloat[95-percentile $\Delta E_{ab}^*$ color error]{\includegraphics[width=0.48\textwidth]{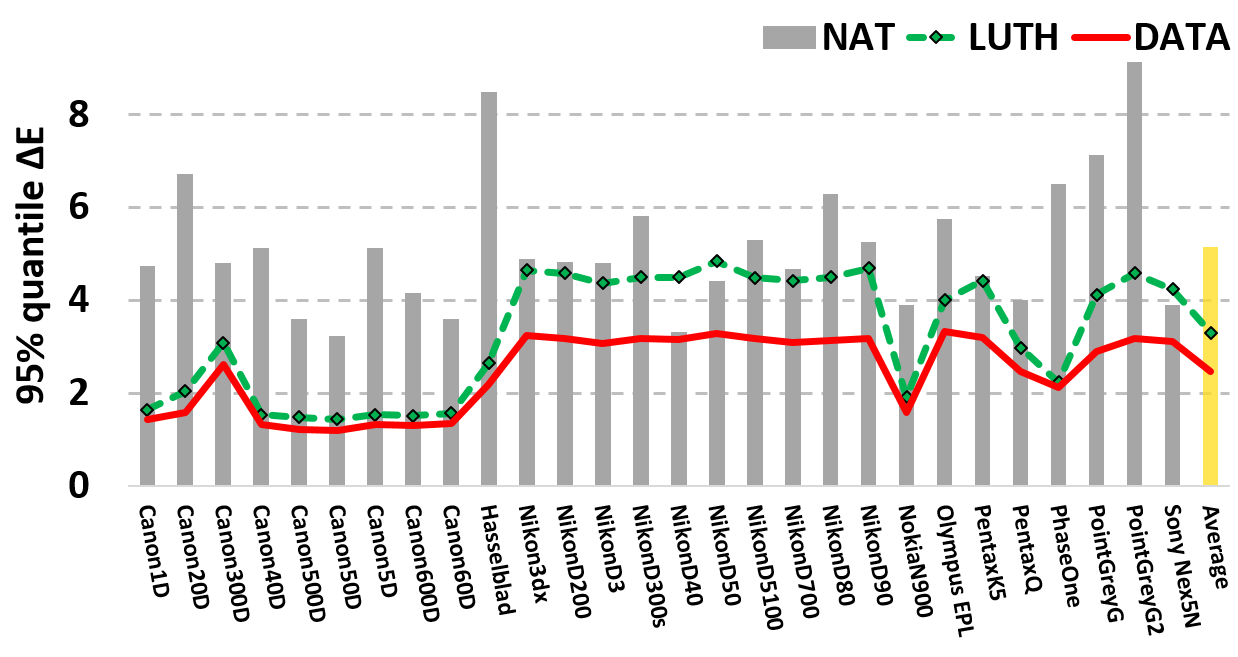} \label{fig:de_95p_28cam} }
    \caption{Mean (a)  and 95-percentile (b) $\Delta E_{ab}^*$ errors for 28 cameras.  The grey-bars show the color errors for {\bf NAT}ive color correction. The dashed green lines with black circles show the results by using the {\bf LUTH}er-condition optimised filter. The results of the {\bf DATA}-driven optimisation are plotted in solid red lines.}
    \label{fig:de_allmethods_28cam}
\end{figure}

Now, we calculate the mean $\Delta E_{ab}^*$ error (for the 102 lights and 1995 reflectances) for each of 28 cameras~\cite{jiang2013space}. For each camera we calculate the optimal Luther-condition and Data-driven optimal filters (where as before the Luther-condition filter seeds the Data-driven optimisation). Per camera, Figure~\ref{fig:de_allmethods_28cam}
summarises the per camera mean and 95 percentile $\Delta E_{ab}^*$ performance.

Grey bars  in Figs. 4a and 4b show respectively the mean and 95 percentile error performance of native (un-filtered) color corrected RGBs for the 28 cameras. Respectively, the dashed green and solid red lines record the performance of the best Luther-condition and  Data-driven filters. 

It is evident that the optimised filters support improved color measurement for all 28 cameras and on average the performance increment is significant. For many cameras the Data-driven optimised filter delivers significantly better color measurement performance compared with using the  Luther-condition optimised filter.

\begin{table}[!t]
\centering
\caption{$\Delta E_{ab}^*$ statistics of the color corrected  {\bf NAT}ive camera, the color corrected camera with the {\bf LUTH}er-condition optimised filter, and  the color corrected camera with the {\bf DATA}-driven optimised filter for Canon 5D Mark II camera under different lighting cases}
\renewcommand{\arraystretch}{1.3}
\setlength{\arrayrulewidth}{0.5mm}
\setlength\tabcolsep{6pt} 
\begin{tabular}{|l|c|c|c|c|c|c|}
\hline
&Mean & Median & 90\% & 95\% & 99\% & Max \\ \hline
\multicolumn{7}{|c|}{CIE D65}     \\ \hline
\textbf{NAT}  &1.65	&1.03	&3.55	&4.94	&11.23	&19.29  \\ \hline
\textbf{LUTH}&0.46	&0.25	&1.09	&1.45	&3.49	&5.90 \\ \hline
\textbf{DATA} & 0.38    &0.20    &0.93  & 1.25   &2.45  & 4.62  \\ \hline
\multicolumn{7}{|c|}{CIE A}   \\\hline
\textbf{NAT}   &2.30	&1.44	&4.65	&6.17	&16.96	&26.41 \\ \hline
\textbf{LUTH}   & 0.64	&0.40 &1.33	&1.84	&4.75	&8.19 \\\hline
\textbf{DATA} & 0.44    &0.26           &1.02          & 1.41    &2.81 &4.31 \\ \hline
\multicolumn{7}{|c|}{102 illuminants}   \\ \hline
\textbf{NAT}  & 1.72   & 1.02 & 3.68 & 5.12 & 12.94 & 28.39 \\ \hline
\textbf{LUTH}      &0.53   &0.30 & 1.15 & 1.65 & 4.11  &6.83  \\ \hline
\textbf{DATA} & 0.41   &0.21  &0.96  & 1.32         &2.78 & 6.78  \\ \hline
\end{tabular}
\label{tab:Canon_de_all}
\end{table}

\subsection{Smooth and Bounded Transmittance Filters} 

\begin{figure}[!tb]%
    \centering
    \subfloat[no basis]{\includegraphics[width=0.4\textwidth]{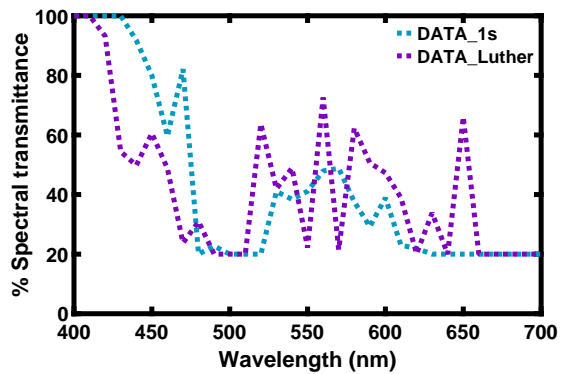} \label{fig:no_basis} }	\hfill
	 \subfloat[6-basis series]{\includegraphics[width=0.4\textwidth]{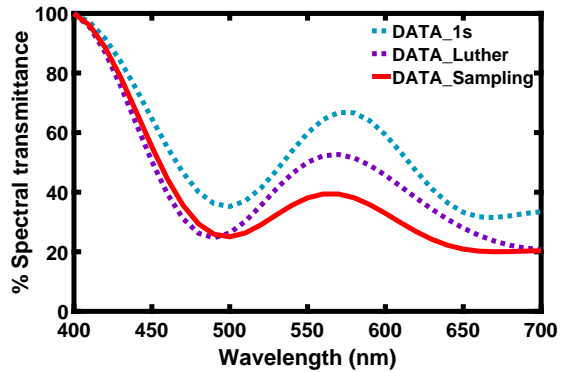} \label{fig:6cosine} } \hfill
 	 \subfloat[8-basis series]{\includegraphics[width=0.4\textwidth]{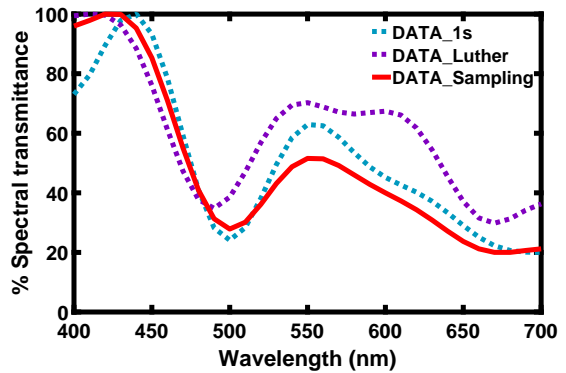} \label{fig:8cosine} }\hfill
    \caption{Spectral transmittance of Data-driven optimised filters for the Canon 5D Mark II camera. In (a), the filters are solved with no smoothness constraint, (b) and (c) constrain the filters to be constructed by 6- and 8- Cosine basis respectively. Dotted blue lines (`{\bf DATA\_1s}') show the filters obtained when the optimisation is initialised with the 100\% transmittance filter, dotted purple lines (`{\bf DATA\_Luther}') initialised with the  \textbf{LUTH}er-condition filter, solid red lines (`{\bf DATA\_Sampling}') with a sample set. All filters are constrained by minimum transmittance of 20\%. }
    \label{fig:reconstructfilters}
\end{figure}%

Both the Luther-condition and Data-driven filters 
absorb much of the available light (low transmittance values) and are far from being smooth, e.g.\ see the derived Luther-condition filter in Fig. 1e. Here across much of the visible spectrum the filter transmits little (below 20\%) of the available light. 
When a strongly light-absorptive filter is placed in front of a camera then we need to either increase the   exposure time (or widen the  aperture) or apply a  higher ISO (which can increase the conspicuity of noise) to obtain an image with the same range of brightnesses (as when a filter is not used). Plus the filter in Fig. 1e is not smooth it may be difficult to manufacture.

In this subsection we will constrain our optimisation so that the calculated filters are smooth and transmit, per wavelength, a minimum amount of incident light (here, 20\%). We enforce smoothness indirectly by assuming that our filters lie within the span of either a 6- or 8-dimensional Cosine basis.

Let us visualise the 20\%  bounded transmittance constraint using the  Canon 5D Mark II camera sensitivities and our Data-driven optimisation. First, we initialise the optimisation (Algorithm 2) using  the uniform vector $\underline{1}$ (100\% transmittance across the spectrum) as the {\it seed}  filter. The optimisation returns the filter, denoted  {\bf DATA\_1s} and is plotted in the blue-dotted line in Fig. 5a. The Data-driven optimisation seeded with the Luther-condition filter {\bf DATA\_Luther}  is also shown in Fig. 5a (dotted purple line).

We repeat this experiment where we both require the derived filters to transmit at least 20\% of the light and also that they belong to the span  the first 6- or first 8-terms in a Cosine basis expansion. When the recovered filter is constrained to lie in the span of a 6-dimensional Cosine basis, the recovered {\bf DATA\_1s} and {\bf DATA\_Luther} - for the two initialisation conditions -  filters are shown in Fig. 5b (respectively blue and purple dotted lines). See Fig. 5c for the filters calculated using an 8-dimensional Cosine basis. 

The red lines shown in Figs. 5b and 5c are the filters optimised by our sampling Algorithm (using Algorithm 3 to find a set of filters to seed algorithm 2 and then choosing the best one that has the best overall performance). The experiment for deriving these filters are described in the next subsection.

By examining Figs. 5a, 5b and 5c it is evident that - when no basis, a 6- or an 8-dimensional Cosine basis are used - that changing the initialisation condition results in a different filter being found.

\subsection{Sampled Optimisation}

Using Algorithm 3, let us run a sampled optimisation. That is for a given Cosine basis we calculate a set of candidate solutions, the sample set ${\cal F}$. Here
we populate $\cal F$ with  20,000 uniformly and randomly generated filters where the angular threshold between any two filters in the set is 1 degree, $\theta = \ang{1}$, see step 7 in Algorithm 3. Each filter in $\cal F$ transmits at least 20\% of the light (and $\cal F$
is populated by filters described as linear combination of a 6- or 8-dimensional Cosine basis).

Each filter in ${\cal F}$ is used to initialise the Data-driven algorithm. That is we find 20,000 optimised filters. The color measurement performance of each filter in this solution set can be calculated. Then we simply choose the filter that delivers the best overall measurement performance. In Figs. 5b and 5c we show the best  sample-optimised filters (red lines)  which respectively lie in the span of a 6- and 8-dimensional Cosine basis (and transmit at least 20\% of the light). Here 'best' is defined to be the filter that results in the smallest mean $\Delta E^{*}_{ab}$ performance.

Table~\ref{tab:reconstructed} reports the $\Delta E_{ab}^*$ color error performance 1995 reflectances and 102 lights~\cite{barnard2002data} for the Canon 5D Mark II sensitivities. The row {\bf NAT} reports the baseline color correction results when a per illuminant based linear correction matrix is applied while no filter is used (note that row 7 in Tables I and row  1 in Table II are the same).

In Table II we report the correction performances in 3 tranches. Rows 2 and 3 correspond to the two filters without using Cosine basis as shown in Fig. 5a. Here we find the best filters with only the 20\% minimum transmittance bound. 
Rows, 4,5 and 6 report the performance when the 3 filters shown in Fig. 5b are used, where the filter is additionally constrained to be in the span of the 6-dimensional Cosine basis. 
Finally, when the filter is constrained to belong to an 8-dimensional Cosine basis, the 3 derived filters lead to the error statistics shown in rows 7 through 9.

Table I reported the color measurement performance of the filters found using an unconstrained optimisation. Table II reports the color measurement results that are found when filters are constrained to have a bounded transmittance (here at least 20\% of the light) and be smooth.
Let us consider the bounded transmittance first. 
Comparing row 9 of Table I to row 3 of Table II we see that adding a lower transmittance bound returns a filter that delivers poorer measurement performance (but still much better compared with the native camera response). 
Additionally, requiring that our filters smooth on top of the minimum also yields relatively poorer performance compared to the unconstrained filter optimisation.

However, with either the 6- and 8-dimensional Cosine basis constraint we can find the best filter by seeding Algorithm 2 with many possible filter initialisations (and then choosing the best filter overall). Here, we find that comparable performance is possible. Compare rows 6 and 9 of Table II to row 9 of Table I. It is remarkable how well a constrained filter can work: the performance is ever so slightly worse than the unconstrained optimisation. But, the filter is much smoother and more likely to be able to be manufactured.

\begin{table}[!th]
\centering
\caption{
$\Delta E_{ab}^*$ statistics of the color corrected  {\bf NAT}ive camera, the color corrected camera with the {\bf DATA}-driven optimised filter solutions obtained when initialised with uniform vector of {\bf 1s}, {\bf Luther}-condition filter and {\bf Sampling} filter set respectively under different constraints for Canon 5D Mark II camera}
\renewcommand{\arraystretch}{1.3}
\setlength{\arrayrulewidth}{0.5mm}
\setlength\tabcolsep{4pt} 
\begin{tabular}{|l|c|c|c|c|c|c|}
\hline
&Mean    & Median & 90\% & 95\% & 99\% & Max    \\\hline
\textbf{NAT}  &1.72	&1.03	&3.68	&5.12	&12.94	&28.39  \\ \hline
\multicolumn{7}{|c|}{minimum transmittance of 20\%} \\ \hline
\textbf{DATA\_1s} & 0.69 & 0.42 & 1.47 & 2.11  & 4.69 & 19.48\\ \hline
\textbf{DATA\_Luther} & 0.58   &0.38  &1.36  & 1.80         &\bf 2.77 &\bf 5.75  \\ \hline
\multicolumn{7}{|c|}{6 cosine basis with 20\% minimum transmittance}  \\ \hline
\textbf{DATA\_1s} & 0.81 &0.49  &1.80  & 2.54   &5.21  & 18.85  \\ \hline
\textbf{DATA\_Luther} &0.94  &0.54  &2.03  & 2.84   &7.00  & 21.14  \\ \hline
\textbf{DATA\_Sampling}  & 0.59 & 0.35  & 1.30  &1.83   & 3.77  & 14.19  \\ \hline
\multicolumn{7}{|c|}{8 cosine basis with 20\% minimum transmittance}   \\ \hline
\textbf{DATA\_1s} & 0.71 &0.38  &1.60  & 2.38   &5.42  & 19.25  \\ \hline
\textbf{DATA\_Luther} &0.62  &0.38  &1.41  & 2.01   &3.47  & 9.52  \\ \hline
\textbf{DATA\_Sampling} &\bf 0.45 &\bf 0.25	&\bf 1.02	&\bf 1.41 & 3.10 &10.63 \\ \hline
\end{tabular}
\label{tab:reconstructed}
\end{table}

\subsection{Sampling vs Optimisation}

It is worth reflecting on our sample-based optimisation. Clearly, that sampling makes such a difference to the performance that our optimisation can deliver (for filtered color measurement) teaches us that the minimisation at hand has many local minima. By sampling we are effectively allowing our minimiser (Algorithm 2) to find many solutions and then we have the latitude to choose the (closer to) global minimum. Given we seed our optimisation with 20,000 filters we might wonder whether we
 need to actually carry out the Data-driven optimisation.
 
In answering this question, first we remark that it is well known that as the dimension of a space increases it is more {\it sparse}. On the Cartesian plane if we have more than 360 vectors (anchored at the origin) then the closest angular distance to at least one vector's nearest neighbours must be less than 1 degree. In 3-dimensions we can have thousands of vectors where every vector is more than 1 degree from its nearest neighbour. 

For our 20,000 member sample  set $\cal F$ we calculated the average angular distance for each element to its nearest neighbour in the set. When $\cal F$ is calculated subject to the 6-dimensional Cosine basis constraint, the average nearest-neighbour distance was 2.6 degrees (with a maximum of 7) and for the 8-dimensional case it was 4.6 degrees (with a maximum of 10). Running the optimisation, Algorithm 2, with each element in $\cal F$, we effectively {\it refine} the initial guess. And, the refinement (difference between the starting and endpoint filter) is on the same order as the average nearest-neighbour distance. 

Significantly, running the optimisation - carrying out the refinement - results in a significant performance increment compared to using the only sample filters. That is we cannot use the sampling strategy alone to find the best optimised filter. The importance of the refinement step will increase as a greater number of basis functions are used in the optimisation.

\section{Conclusion}
In this paper, we developed two algorithms that design transmittance filters which, when placed in front of a camera, make the camera more colorimetric. Our first algorithm is driven by the camera sensitivities themselves. It is well known that a camera that has sensitivities that are a linear transform from the XYZ color matching functions - the camera meets the {\it Luther condition - }can be used to measure color without error. Our first algorithm finds the filter that best satisfies the Luther condition. A second algorithm that tackles color correction for a given set of measured lights and surfaces, which we call Data-driven filter optimisation, is also developed. Both Luther- and Data-driven Filters provide a step change in how well a camera can measure color.

Our default optimisation - though compellingly simple to formulate - deliver filters which are not smooth (difficult to make) and may also transmit very little light. Our optimisations are reformulated to incorporate both smoothness and a lower bound on how much light must be transmitted. Initially, when these constraints are adopted, the solved-for filters work less well. However, experiments demonstrated that our optimisations were highly dependent on the initialisation parameters, specifically the {\it seed} filter (initial guess) that drives the filter evolution. A simple sampling strategy - i.e. severally running the optimisation for a set of judiciously chosen seed filters - allows us to mitigate this problem. Significantly a smooth filter that transmits more than 20\% of the light across the visible spectrum delivers almost as good performance as a very non-transmittive and non-smooth filter (found via the unconstrained optimisations).

\appendix[Implementation]

For  both Algorithms 1 and 2 presented in Section~\ref{sec:method} the filter and the color correction matrices can be found using simple least-squares regression. To remind the reader, given  $A$ and $B$ - $m\times n$ matrices of rank $n$ where $m \geq n$, then the least-squares regression $M$ - an $n\times n$ matrix, mapping $A$ to $B$ ($AM\approx B$)
can be found in closed-form using the Moore-Penrose inverse~\cite{golub2012matrix}:
$$
M=[A^TA]^{-1}A^TB=A^+B
$$

\subsection{Algorithm 1: ALS for the Luther-condition Optimisation}

In step 4 of the algorithm, the optimal filter is found by finding scalars that maps each row of $Q^{i-1}$ to the corresponding row of $\text{\it X}$. The best scalar $\alpha$ mapping the vector  $\underline{v}=[Q^{i-1}_j]^T$ to $
\underline{w}=[\text{\it X}_j]^T$ (for the $j${th} row of the data matrices) can be written in closed form using the Moore-Penrose inverse: 
$\alpha=\frac{\underline{v}^T\underline{w}}{\underline{v}^T\underline{v}}$. 

Similarly, in step 5, the Moore-Penrose inverse can be used for finding $M$. Denoting ${\cal Q}=diag(\underline{f}^i)Q^{i-1}$ then $M^i={\cal Q}^+\text{\it X}= 
[{\cal Q}^T{\cal Q}]^{-1}{
\cal Q}^T\text{\it X}$.

\subsection{Algorithm 2: ALS for the Data-driven Optimisation}

In step~\ref{codeM2}, each $M^i_j$ can be solved directly using the Moore-Penrose inverse. Denoting ${\cal Q}=C_j^T Q^{i-1}_j$ then $M^i_j={\cal Q}^+C_k^T\text{\it X}=[{\cal Q}^T{\cal Q}]^{-1}{
\cal Q}^TC^T_k\text{\it X}$.

In step~\ref{codeD2} of the Data-driven optimisation, the filter $\underline{f}$ is embedded in the equation and so we cannot solve for it directly as we could for the Luther-condition case.

To solve for the filter it is useful to vectorise the minimisation. We recapitulate the minimisation statement of step~\ref{codeD2}:
\begin{equation}
    \min\limits_{\underline{f}} \sum_{j=1}^{cnt} \parallel{(C^T_j diag(\underline{f}) Q_jM_j - C^T_k \text{\it X}})\parallel_{F}^2
    \label{eq:Avec}
\end{equation}

\noindent
The meaning of $\parallel \, \parallel^2_F$ is the Frobenius norm squared, i.e.\ the sum of all the argument terms squared. This Frobenius norm is generally applied to matrices (as here) but can equally be applied to vectors (where the $vec()$ operator stacks the columns of a matrix on top of each other):
\begin{equation}
 \min\limits_{\underline{f}} \sum_{j=1}^{cnt} \parallel{vec(C^T_j diag(\underline{f}) Q_jM_j) - vec(C^T_k \text{\it X})}\parallel_{F}^2
 \label{eq:Amin}
\end{equation}

Now let us rewrite the diagonal filter matrix as a summation of each value in the diagonal, $f_i$, multiplied with a single entry matrix $D_i$ as $diag(\underline{f}) = \sum_{i = 1}^{31} f_i D_i$. Here, $D_i$ is a $31 \times 31$ matrix with a single non-zero entry at $D(i,i) = 1$. 
By substituting this new filter representation into the first term of the minimisation in Equation~(\ref{eq:Amin}), we obtain
\begin{equation}
vec(C^{T}_jdiag(\underline{f})Q_jM_j) = \sum\limits_{i=1}^{31}f_i \, vec( C^{T}_jD_{i}Q_jM_j)\\
	\label{eq:vecdivided}
\end{equation}

Now let us denote a matrix $V_j = [\underline{v}_{1,j} \, \underline{v}_{2,j} \, \cdots \, \underline{v}_{31,j}]$ where its column represents $\underline{v}_{i,j}=vec(C^{T}_jD_{i}Q_jM_j)$, Equation~(\ref{eq:vecdivided}) can be expressed more compactly as
\begin{equation}
    vec(C^{T}_jdiag(\underline{f})Q_jM_j) = V_j\underline{f}
\end{equation}
where $\underline{f} = [f_1 \; f_2\; ...\; f_{31}]^T$. Note that if  $C_j$ is a $31 \times n$ matrix then $C^{T}_jD_iQ_jM_j$ is an $n \times 3$ matrix (where $3$ denotes the number of color channels) and thus $\underline{v}_{i,j}$ is $3n \times 1$ which makes matrix $V_j$ have size of $3n \times 31$. 

Now we stack all $V_j$, $j = 1,2,..,cnt$  matrices  (under {\it cnt} different lighting conditions) on top of each other making an $(3n*cnt)\times 31$ matrix, $V$. Similarly we stack all $cnt$ targeting XYZs on top of each other denoted as $\underline{w} = vec(C_k^T X)$ which has the size of $(3n*cnt)\times 1$. We remind the reader that $k$ might equal $j$. Or, $k$ might denote a single privileged illuminant such as CIE D65. 

Now the minimisation in Equation~(\ref{eq:Amin}) can be equivalently rewritten as:

\begin{equation}
    \min \limits_{\underline{f}} \parallel V\underline{f} - \underline{w} \parallel^2_F
    \label{eq:reform_pb}
\end{equation}
The best $\underline{f}$ can be found in closed form using the Moore-Penrose inverse:
\begin{equation}
    \underline{f}= (V)^{+}\underline{w}=[V^TV]^{-1}V^T\underline{w}.
    \label{eq:reform_pb2}
\end{equation}

\subsection{Filter Constraints}
Equation (\ref{eq:reform_pb2}) solves for the 31-component $\underline{f}$ in one step. Suppose we write  $\underline{f}=B\underline{c}$. We constrain the filter to be describable by a linear basis ($B$ is $31\times m$ where $1\leq m \leq 31$).  Additionally, the filter is restrained by a minimum $f_{min}$ and maximum $f_{max}$ bounds on the transmittance.
Then to solve for the filter we find the coefficient vector $\underline{c}$ that minimises:
\begin{equation}
    \min \limits_{\underline{c}} \parallel VB\underline{c} - \underline{w} \parallel^2_F
    \;\text{s.t.} \; f_{min}\leq{B\underline{c}}\leq f_{max}
     \label{eq:reform_pb4}
\end{equation}
Equation (\ref{eq:reform_pb4}) where there is a quadratic objective function and linear inequality constraints can be solved using quadratic programming~\cite {luenberger1984linear}.

\section*{Acknowledgment}
This work was supported by EPSRC under Grant EP/S028730. The authors would also like to thank Dr. Javier Vazquez-Corral for his insightful comments. 

\ifCLASSOPTIONcaptionsoff
  \newpage
\fi

\bibliographystyle{IEEEtran}
\bibliography{IEEEabrv,main.bbl}

\end{document}